\documentclass[sigconf,oneside,nonacm]{acmart}
\renewcommand\footnotetextcopyrightpermission[1]{}

\AtBeginDocument{%
  }

\setcopyright{acmlicensed}
\copyrightyear{2026}
\acmYear{2026}
\acmConference[ACM MM]{the 34th ACM International Conference on Multimedia}{10–14 November 2026}{Rio de Janeiro, Brazil}
\acmISBN{978-1-4503-XXXX-X/2018/06}

\usepackage{multirow} 
\usepackage{amsmath}
\usepackage{balance}
\newcommand{\best}[1]{\textbf{#1}}

\begin{document}

\title{Rethinking Video Human–Object Interaction: Set Prediction over Time for Unified Detection and Anticipation}


\author{Yuanhao Luo}
\authornote{Both authors contributed equally to this research.}
\email{yuanhao.luo@student.kit.edu}
\affiliation{%
  \institution{Karlsruhe Institute of Technology}
  \city{Karlsruhe}
  \country{Germany}
}

\author{Di Wen}
\authornotemark[1]
\email{di.wen@kit.edu}
\affiliation{%
  \institution{Karlsruhe Institute of Technology}
  \city{Karlsruhe}
  \country{Germany}
}

\author{Kunyu Peng}
\authornote{Corresponding author: \texttt{kunyu.peng@kit.edu}}
\email{kunyu.peng@kit.edu}
\affiliation{%
  \institution{Karlsruhe Institute of Technology}
  \city{Karlsruhe}
  \country{Germany}
}
\affiliation{%
  \institution{INSAIT, Sofia University}
  \city{Sofia}
  \country{Bulgaria}
}

\author{Ruiping Liu}
\affiliation{%
  \institution{Karlsruhe Institute of Technology}
  \city{Karlsruhe}
  \country{Germany}
}

\author{Junwei Zheng}
\affiliation{%
  \institution{Karlsruhe Institute of Technology}
  \city{Karlsruhe}
  \country{Germany}
}
\affiliation{%
  \institution{ETH Zurich}
  \city{Zurich}
  \country{Switzerland}
}

\author{Yufan Chen}
\affiliation{%
  \institution{Karlsruhe Institute of Technology}
  \city{Karlsruhe}
  \country{Germany}
}

\author{Jiale Wei}
\affiliation{%
  \institution{Karlsruhe Institute of Technology}
  \city{Karlsruhe}
  \country{Germany}
}

\author{Rainer Stiefelhagen}
\affiliation{%
  \institution{Karlsruhe Institute of Technology}
  \city{Karlsruhe}
  \country{Germany}
}

\renewcommand{\shortauthors}{Luo and Wen, et al.}

\begin{teaserfigure}
  \centering
  \includegraphics[width=\textwidth]{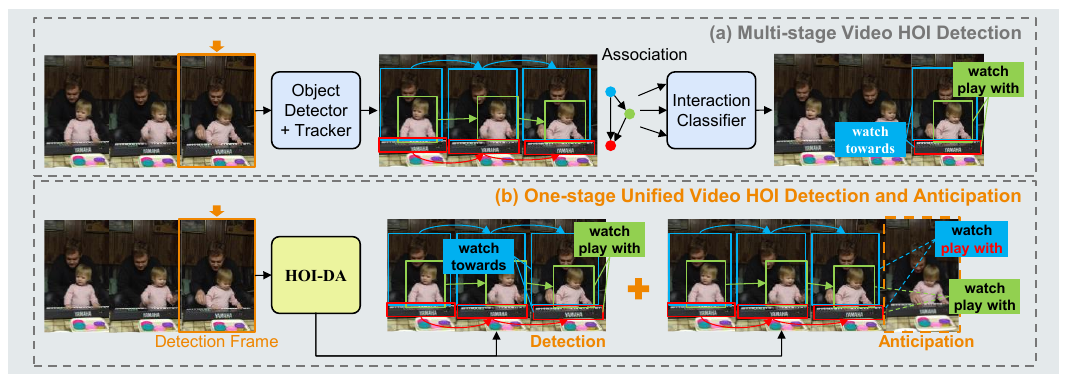}
  \vskip-3ex
  \caption{
  Comparison between conventional multi-stage video HOI pipelines and HOI-DA.
  Unlike prior methods that separate detection, association, and interaction reasoning, HOI-DA uses shared pair hypotheses to jointly predict present interactions and future HOIs from the observed clip.
  }
  \Description{The figure contrasts conventional multi-stage video HOI pipelines with the proposed unified formulation. 
  The top row illustrates a typical two-stage pipeline, where human and object instances are first detected and tracked, followed by post-hoc association and interaction classification. 
  In this design, interaction reasoning operates on externally constructed pairs, and future prediction is treated as a separate downstream task. 
  The bottom row shows HOI-DA, a one-stage pair-centric framework that directly models a set of persistent human--object pair hypotheses from the observed clip. 
  These pair representations are maintained across time and are used to jointly predict present interactions and future HOIs without explicit tracking or association. 
  Unlike multi-stage approaches, the same pair identity is preserved from detection to anticipation, enabling temporally consistent reasoning over interaction evolution. 
  This unified formulation allows anticipation to be learned as a structured extension of current interaction states, rather than as an independent prediction task.
}
  \label{fig:archi}
  \vskip-0.5ex
\end{teaserfigure}

\begin{abstract}

Video-based human--object interaction (HOI) understanding requires both detecting ongoing interactions and anticipating their future evolution. However, existing methods usually treat anticipation as a downstream forecasting task built on externally constructed human--object pairs, limiting joint reasoning between detection and prediction. In addition, sparse keyframe annotations in current benchmarks can temporally misalign nominal future labels from actual future dynamics, reducing the reliability of anticipation evaluation. To address these issues, we introduce \textbf{DETAnt-HOI}, a temporally corrected benchmark derived from VidHOI and Action Genome for more faithful multi-horizon evaluation, and HOI-DA, a pair-centric framework that jointly performs subject--object localization, present HOI detection, and future anticipation by modeling future interactions as residual transitions from current pair states. Experiments show consistent improvements in both detection and anticipation, with larger gains at longer horizons. Our results highlight that anticipation is most effective when learned jointly with detection as a structural constraint on pair-level video representation learning. Code will be publicly available.

\end{abstract}

\ccsdesc[500]{Computing methodologies~Activity recognition and understanding}
\keywords{Activity Recognition and Understanding, Video Understanding}


\maketitle

\section{Introduction}
\label{sec:intro}

Video-based Human--Object Interaction (HOI) understanding has advanced from static-image HOI detection toward spatio-temporal reasoning over who interacts with what, how, and when~\cite{girdhar2019video,tu2022tutor}. Yet many practical scenarios, such as collaborative robotics~\cite{mascaro2023hoi4abot}, proactive safety monitoring, and anticipatory human--machine interaction, require more than recognizing ongoing interactions: they require predicting how already-observed human--object pairs will evolve over future time horizons. The central challenge is therefore not prediction alone, but \emph{temporal consistency}: the representation that grounds a present interaction should also remain valid for reasoning about its future evolution.

Despite this need, joint video HOI detection and anticipation remain relatively underexplored~\cite{ni2023gaze}. Existing anticipation-oriented methods mostly follow a two-stage formulation: they first detect or track instances, then classify or forecast interactions over pre-constructed human--object pairs~\cite{ni2023gaze,mascaro2023hoi4abot}. Such a design can produce competitive pipelines, but it makes anticipation a downstream prediction problem over externally formed candidates rather than a structured evolution of the same pair representation. As a result, the model is only weakly constrained to learn how an interaction persists, changes, or dissolves over time. In other words, the dominant difficulty is not simply future prediction, but whether pair identity and pair state are preserved as first-class objects throughout the model.

A second limitation lies in evaluation. Widely used benchmarks such as VidHOI~\cite{chiou2021vidhoi} and Action Genome~\cite{ji2020action} rely on sparse keyframe annotations rather than fully temporally continuous supervision. Under such protocols, nominal ``future'' labels may be separated from the observed clip by long or irregular temporal gaps, so reported anticipation performance can be influenced by annotation sparsity in addition to genuine future dynamics. This weakens the reliability of evaluation and makes it harder to determine whether a model is truly learning temporally grounded anticipation.

We address both issues with a simple unifying principle:
\emph{Future interaction states should be modeled as structured evolutions of the current pair state}.
Rather than predicting future HOIs independently, we represent them as residual transitions from present pair representations. This leads to \textbf{HOI-DA}, a unified pair-centric architecture that jointly performs pair localization, present HOI detection, and multi-horizon anticipation within a shared representation space. Under this view, anticipation is not a downstream add-on to detection, but a structural constraint on how pair-level interactions are represented over time.

Realizing this principle requires more than naive parameter sharing. If future reasoning is learned in the same feature space without additional structure, predictions can collapse toward the current state or become redundant across horizons. To address this, we introduce \emph{dual orthogonality regularization}, which separates present interaction grounding from future interaction change and enforces non-redundant temporal structure across different anticipation horizons. To further improve robustness under long-tailed HOI distributions, we incorporate a \emph{language-guided semantic branch} that injects vocabulary-level semantic structure into both present detection and future prediction.

To evaluate anticipation under temporally meaningful conditions, we further establish \textbf{DETAnt-HOI}, a corrected benchmark built from VidHOI~\cite{chiou2021vidhoi} and Action Genome~\cite{ji2020action}. DETAnt-HOI enforces temporal continuity through supplementary annotation and controlled clip construction, so that future labels correspond more faithfully to actual future dynamics rather than annotation artifacts.

\noindent
Our contributions are threefold:
\begin{itemize}
    \item \textbf{A unified formulation of HOI detection And anticipation.}
    We propose HOI-DA, a pair-centric architecture that models future interactions as residual transitions from present pair states, enabling temporally consistent reasoning across detection and anticipation.

    \item \textbf{Dual orthogonality regularization for structured temporal reasoning.}
    We introduce constraints that separate present grounding from future change and enforce diversity across anticipation horizons.

    \item \textbf{A temporally corrected benchmark for unified HOI detection and anticipation.}
    We establish DETAnt-HOI, which reduces annotation-induced temporal discontinuities and enables more reliable evaluation of multi-horizon anticipation.
\end{itemize}

\section{Related Work}
\label{sec:related}

\noindent\textbf{HOI Detection in Images and Videos.}
HOI detection aims to produce $\langle$human, verb, object$\rangle$ triplets from visual input.
Early approaches rely on two-stage pipelines that first detect humans and objects and then classify interactions over enumerated pairs~\cite{Chao_2018_WACV,Gkioxari_2018_CVPR,gao2018ican}.
More recent work adopts one-stage, query-based formulations built on set prediction~\cite{carion2020detr,kim2021hotr,tamura2021qpic,zhang2021mining}, further improved by multi-scale modeling, relational context, pose-aware reasoning, and structural priors~\cite{kim2022mstr,kim2023relational,park2023viplo,ma2023fgahoi,yang2025no,li2026dqen}.
Despite these advances, image HOI assumes that interactions are fully observable in a single frame.

In videos, the core challenge shifts from pair construction to pair persistence across time.
Existing methods introduce temporal reasoning via trajectories, graphs, tubelets, or prompt-based modeling~\cite{chiou2021vidhoi,wang2021spatio,tu2022tutor,xi2023open,wang2024interaction,gu2025hoi,wu2025hiergat}, yet pair identity is typically recovered through post-hoc association rather than enforced as a first-class object.
As a result, temporal modeling operates on externally constructed pair candidates rather than on a unified pair representation, limiting the ability to model how the same interaction instance evolves over time.
In contrast, HOI-DA treats pair persistence as an architectural primitive, maintaining identity within the model through time-aligned pair slots.

\noindent\textbf{Action Anticipation and HOI Forecasting.}
Anticipating future human behavior has been widely studied in egocentric video, focusing on predicting actions, objects, or attention signals from partial observations~\cite{liu2020forecasting,thakur2024leveraging,grauman2022ego4d}.
In third-person video, HOI anticipation remains comparatively underexplored~\cite{ni2023gaze}.
Existing approaches either decouple detection and prediction~\cite{ni2023gaze,mascaro2023hoi4abot} or operate at a coarse semantic level using language models without spatial grounding~\cite{zhao2023antgpt,kim2024palm}.
Consequently, current methods lack either pair-level grounding or temporally controlled multi-horizon supervision, and do not model temporally persistent pair evolution.
HOI-DA addresses both limitations by jointly modeling detection and anticipation over persistent pair representations, while DETAnt-HOI enforces temporally consistent evaluation.

\noindent\textbf{Vision--Language Priors for HOI.}
Due to the long-tail distribution of verbs, many works incorporate language priors to improve HOI recognition.
Representative approaches leverage CLIP-based embeddings, pretrained relation-aware representations, or multimodal prompting to enhance interaction classification~\cite{liao2022genvlkt,yuan2022rlip,yuan2023rlipv2,ning2023hoiclip,mao2023clip4hoi,cao2023detecting,yang2024open,lei2024exploring,lei2024ez}.
More recent work further explores diffusion models and large multimodal models for HOI reasoning~\cite{yang2023diffhoi,kang2024vlm,xuan2026zero}.
However, these methods are designed for current-frame interaction understanding.
Our setting introduces an additional requirement: semantic priors must remain causally valid for future anticipation, i.e., independent of unobserved future frames.
HOI-DA satisfies this constraint by integrating language priors into a unified pair-centric model that regularizes both present detection and future prediction.

Taken together, existing work improves interaction recognition, temporal modeling, and semantic priors largely in isolation, without enforcing a representation that preserves pair identity and remains predictive across future horizons.

\section{Methodology}
\label{sec:method}

\subsection{Problem Formulation}
\label{sec:setup}

Given an observed video clip of $L$ frames, our goal is to jointly perform
(i)~HOI detection within the observed window and
(ii)~HOI anticipation at future horizons
$\mathcal{H}=\{h_1,\dots,h_{|\mathcal{H}|}\}$.
An HOI instance is defined as a triplet
$\langle \mathrm{human},\,\mathrm{verb},\,\mathrm{object} \rangle$.

Within the observed clip, the model predicts subject--object localization and present HOI labels for a set of persistent pair hypotheses.
Beyond the observation window, it forecasts future verb labels at time $L+h$, $h\in\mathcal{H}$, for the \emph{same} pair hypotheses established from observed evidence; future bounding boxes are not predicted.

Let $\mathcal{Y}_{1:L}$ denote the HOI annotations in the observed clip, and
$\mathcal{Y}_{L+h}$ those at future horizon $h$.
Our objective is to learn a representation that remains grounded enough for present detection while preserving temporally transferable structure for future forecasting.
The key challenge is therefore not only to predict future verbs, but to maintain
\emph{pair-level semantic continuity} across present grounding and future reasoning.

Our design principle is simple:
\emph{Future interaction states should be modeled as structured evolutions of the current pair state}.
This motivates a unified pair-centric architecture in which anticipation is not appended after detection, but constrains the 
pair representations. 

\subsection{Architecture Overview}
\label{sec:overview}

\begin{figure*}[t]
  \centering
  \includegraphics[width=\textwidth]{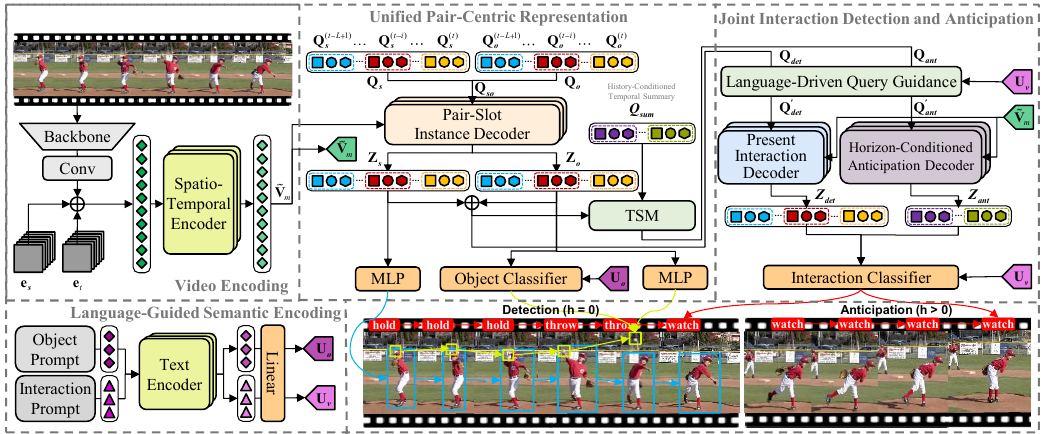}
  \vskip-2ex
  \caption{
    Overview of our model. Given an observed clip, HOI-DA builds a shared spatio-temporal visual memory and uses a unified pair-centric decoder for subject--object localization, present HOI prediction, and multi-horizon future verb anticipation. A language-guided semantic branch regularizes object and verb prediction, while dual residual orthogonality regularization separates future dynamics from present interaction states and across future horizons.}
    \vskip-2ex
    \Description{The figure illustrates the overall architecture of HOI-DA for unified human--object interaction (HOI) detection and anticipation. Given an observed video clip, a visual backbone and spatio-temporal encoder extract a shared video representation, forming a global memory for interaction reasoning. 
    
    A unified pair-centric decoder constructs a fixed set of subject--object pair slots, where pair identity is preserved across time without relying on external detection, tracking, or post-hoc association. These pair representations are used to jointly perform instance localization and present-time HOI prediction through the interaction decoder. 
    
    To enable future anticipation, the Temporal Summary Module aggregates information over the full interaction history and produces horizon-specific queries, which are further processed by the anticipation decoder to predict future verb labels at multiple time horizons. 
    
    In parallel, a language-guided semantic branch injects object and verb priors into the shared representation, improving generalization under long-tail distributions. Dual orthogonality regularization is applied to decouple present interaction grounding from future dynamics and to enforce diversity across different anticipation horizons. 
    
    Overall, the architecture unifies detection and anticipation within a shared pair-centric representation, enabling temporally consistent reasoning over persistent interaction instances.}
  \label{fig:archi}
\end{figure*}

HOI-DA is built around a unified pair-centric representation that jointly supports present HOI
detection and future anticipation.
Instead of first constructing human--object pairs and then attaching a separate forecasting head,
HOI-DA treats pair identity, present interaction state, and future evolution as coupled outputs
of a shared representation space.

A visual backbone~\cite{he2016deep} and Transformer encoder~\cite{vaswani2017attention}
map the observed clip into a spatio-temporal memory
$\mathbf{V}_m \in \mathbb{R}^{P_v \times D}$,
where $P_v$ is the number of visual tokens and $D$ is the hidden dimension.
A unified pair-centric decoder then performs persistent pair construction, present interaction
modeling, history-conditioned temporal summarization, and horizon-specific anticipation.
In parallel, a language-guided semantic branch injects vocabulary-level structure into object and
verb prediction.
The complete architecture is shown in Fig.~\ref{fig:archi}.

\subsection{Unified Pair-Centric Decoder}
\label{sec:decoder}

\noindent\textbf{Pair-Slot Instance Decoder.}
A unified formulation requires the model to reason about the same human--object pair across
time, rather than over independently formed candidates.
We therefore introduce two learnable query tensors
$\mathbf{Q}_s, \mathbf{Q}_o \in \mathbb{R}^{P \times L \times D}$,
where $P$ is the number of persistent pair slots.
For each slot index $i$, the subject and object queries are aligned by construction across time,
yielding a slot space in which pair identity is preserved without external tracking.

The two query streams are updated by stacked self-attention and cross-attention over
$\mathbf{V}_m$, producing subject and object embeddings
$\mathbf{Z}_s, \mathbf{Z}_o \in \mathbb{R}^{P \times L \times D}$.
We then form a shared pair representation by
\begin{equation}
\mathbf{Q}_{\mathrm{pair}} = \mathbf{Z}_s + \mathbf{Z}_o
\in \mathbb{R}^{P \times L \times D},
\label{eq:q_pair}
\end{equation}
where role-specific localization remains attached to the subject and object streams, while
$\mathbf{Q}_{\mathrm{pair}}$ serves as the shared carrier of pair-level semantics.

\noindent\textbf{Present Interaction Decoder.}
To ground the shared pair representation in the observed clip, we add a learnable detection-task
embedding $\mathbf{e}_{\mathrm{task}}^{\mathrm{det}} \in \mathbb{R}^{D}$:
\begin{equation}
\mathbf{Q}_{\mathrm{det}} =
\mathbf{Q}_{\mathrm{pair}} + \mathbf{e}_{\mathrm{task}}^{\mathrm{det}},
\label{eq:q_det}
\end{equation}
and refine it with a dedicated decoder to obtain
$\mathbf{Z}_{\mathrm{det}} \in \mathbb{R}^{P \times L \times D}$.
Its final-frame slice
\begin{equation}
\bar{\mathbf{Z}}_{\mathrm{det}}
\triangleq
\mathbf{Z}_{\mathrm{det}}^{(L)}
\in \mathbb{R}^{P \times D},
\label{eq:z_det_last}
\end{equation}
defines the observation-boundary interaction state that anchors future residual modeling.

\noindent\textbf{Temporal Summary Module.}
Future anticipation should depend on the full observed interaction history rather than direct
extrapolation from a single boundary frame.
For each horizon $h\in\mathcal{H}$, we introduce learnable horizon anchors
$\mathbf{E}^{(h)} \in \mathbb{R}^{P \times D}$ and let them attend over the shared pair
trajectory:
\begin{equation}
\mathbf{Q}_{\mathrm{sum}}^{(h)}
=
\mathrm{CrossAttn}\!\left(
\mathbf{E}^{(h)},\,
\mathbf{Q}_{\mathrm{pair}},\,
\mathbf{Q}_{\mathrm{pair}}
\right)
\in
\mathbb{R}^{P \times D},
\label{eq:temporal_summary}
\end{equation}
where $\mathbf{Q}_{\mathrm{pair}} \in \mathbb{R}^{P \times L \times D}$ is flattened over slot
and time into a sequence of length $PL$.
The resulting anticipation query is therefore history-conditioned rather than copied from the
last frame alone.

\begin{figure}[t]
  \centering
  \includegraphics[width=\columnwidth]{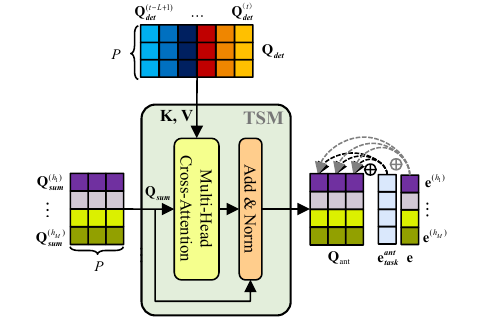}
  \caption{
    \textbf{Temporal Summary Module.}
    Learnable horizon anchors attend over the full shared pair representation
    $\mathbf{Q}_{\mathrm{pair}}$ and produce compact horizon-specific anticipation queries.
  }
  \Description{The figure illustrates the Temporal Summary Module (TSM) used to construct horizon-specific anticipation queries from the shared pair representation. The module takes as input the full temporal sequence of pair representations $\mathbf{Q}_{\mathrm{pair}}$ (serving as keys and values) and a set of learnable horizon anchors $\mathbf{Q}_{\mathrm{sum}}^{(h)}$ for each future horizon. 
  Each horizon anchor attends over the entire observed interaction trajectory through multi-head cross-attention, aggregating information across all time steps rather than relying on the last frame alone. 
  The resulting summarized representation is then combined with task and horizon embeddings to produce the final anticipation query $\mathbf{Q}_{\mathrm{ant}}^{(h)}$. 
  By conditioning future prediction on the full interaction history, the module enables history-aware anticipation and avoids degenerate solutions where future states collapse to the observation boundary. 
  This design allows the model to capture temporally evolving interaction dynamics and generate distinct predictions for different anticipation horizons.}
  \label{fig:tsm}
\end{figure}

\noindent\textbf{Horizon-Conditioned Anticipation Decoder.}
Each temporal summary is conditioned by an anticipation-task embedding
$\mathbf{e}_{\mathrm{task}}^{\mathrm{ant}} \in \mathbb{R}^{D}$ and a horizon-specific embedding
$\mathbf{e}^{(h)} \in \mathbb{R}^{D}$:
\begin{equation}
\mathbf{Q}_{\mathrm{ant}}^{(h)}
=
\mathbf{Q}_{\mathrm{sum}}^{(h)}
+
\mathbf{e}_{\mathrm{task}}^{\mathrm{ant}}
+
\mathbf{e}^{(h)},
\qquad h\in\mathcal{H},
\label{eq:q_ant}
\end{equation}
which is decoded into a horizon-specific anticipation state
$\mathbf{Z}_{\mathrm{ant}}^{(h)} \in \mathbb{R}^{P \times D}$ for future verb prediction at time
$L+h$.

\noindent\textbf{Residual View of Future Dynamics.}
Rather than re-encoding the entire future state at each horizon, we model future anticipation as
a residual departure from the observation-boundary interaction state:
\begin{equation}
\mathbf{R}_i^{(h)}
=
\mathbf{Z}_{\mathrm{ant},i}^{(h)} - \bar{\mathbf{Z}}_{\mathrm{det},i},
\label{eq:residual}
\end{equation}
where $\bar{\mathbf{Z}}_{\mathrm{det},i}\in\mathbb{R}^{D}$ is the $i$-th row of
$\bar{\mathbf{Z}}_{\mathrm{det}}$.
This residual view makes the anticipation branch focus on interaction change rather than
re-describing the present interaction context.

\subsection{Language-Guided Semantic Branch}
\label{sec:text}

To inject semantic structure into the HOI label space, especially under long-tail verb
distributions, we use a lightweight language-guided branch.
A pretrained RoBERTa-based~\cite{liu2019roberta} text encoder maps object names and verb prompts
into projected embeddings
\begin{equation}
\mathbf{U}_o \in \mathbb{R}^{C_o \times D},
\qquad
\mathbf{U}_v \in \mathbb{R}^{C_v \times D},
\label{eq:text_proto}
\end{equation}
where $C_o$ and $C_v$ denote the numbers of object and verb categories.
These embeddings serve two roles: they act as semantic classification prototypes for object and
verb prediction, and they provide an auxiliary guidance source during decoding.
In this way, the branch injects vocabulary-level structure into pair-centric reasoning with
negligible computational overhead while remaining causally valid for future prediction.

\subsection{Training Objective}
\label{sec:loss}

\noindent\textbf{Shared Bipartite Matching.}
Since present detection and future anticipation are defined on the same pair hypotheses, both
tasks share a common slot assignment.
We match the $P$ predicted pair slots to the $M$ ground-truth HOI instances in the observed
clip using bipartite matching~\cite{carion2020detr}:
\begin{equation}
\hat{\omega}
=
\arg\min_{\omega\in\Omega}
\sum_{k=1}^{M}
\mathcal{C}\bigl(
\hat{y}_{\omega(k)},\,y_k
\bigr),
\label{eq:hungarian}
\end{equation}
where $\Omega$ is the set of valid one-to-one assignments and $\mathcal{C}$ combines
localization and classification terms over the observed clip.
The resulting assignment is reused for future supervision at all horizons.
If frame $L+h$ is unavailable, the sample is masked out at horizon $h$; if it is available but the matched pair has no active future verb, the target is an all-zero multi-hot vector and is treated as a valid negative.

\noindent\textbf{Detection and Anticipation Losses.}
We use the same focal-style multi-label verb loss~\cite{lin2017focal} for present and future
verb prediction:
\begin{equation}
\ell_v(\hat{\mathbf{P}}_v,\mathbf{Y}_v)
=
-\frac{1}{\max(N_+,1)}
\begin{aligned}[t]
\sum_{i,c}\Big[
&Y_{ic}(1-\hat{P}_{ic})^2 \log \hat{P}_{ic} \\
&+ (1-Y_{ic})\hat{P}_{ic}^2 \log(1-\hat{P}_{ic})
\Big]
\end{aligned}
\label{eq:verb_focal}
\end{equation}
where $N_+=\sum_{i,c}Y_{ic}$ is the number of positive verb labels.

The present-time detection loss is
\begin{equation}
\mathcal{L}_{\mathrm{det}}
=
\sum_{k\in\{s,o\}}
\Big(
\lambda_b \mathcal{L}_b^k
+
\lambda_{\mathrm{GIoU}} \mathcal{L}_{\mathrm{GIoU}}^k
\Big)
+
\lambda_o \mathcal{L}_{\mathrm{cls}}^o
+
\lambda_v \mathcal{L}_{\mathrm{verb}}^{\mathrm{cur}},
\label{eq:loss_det}
\end{equation}
with
\begin{equation}
\mathcal{L}_{\mathrm{verb}}^{\mathrm{cur}}
=
\ell_v(
\hat{\mathbf{P}}_v^{\mathrm{cur}},
\mathbf{Y}_v^{\mathrm{cur}}
).
\label{eq:loss_verb_cur}
\end{equation}

For each horizon $h\in\mathcal{H}$, the anticipation loss is
\begin{equation}
\mathcal{L}_{\mathrm{verb}}^{(h)}
=
\ell_v\!\left(
\hat{\mathbf{P}}_v^{(h)},
\mathbf{Y}_v^{(h)}
\right),
\label{eq:loss_verb_future}
\end{equation}
and the total anticipation loss is
\begin{equation}
\mathcal{L}_{\mathrm{ant}}
=
\sum_{j=1}^{|\mathcal{H}|}
\lambda_v^{(h_j)}
\mathcal{L}_{\mathrm{verb}}^{(h_j)}.
\label{eq:loss_ant}
\end{equation}
We use a normalized geometric schedule to emphasize near-term forecasting:
\begin{equation}
\lambda_v^{(h_j)}
=
\eta\,
\frac{\gamma^{\,j-1}}
{\sum_{m=1}^{|\mathcal{H}|}\gamma^{\,m-1}},
\qquad
0<\gamma\le 1,
\label{eq:horizon_weight}
\end{equation}
where horizons are ordered from near to far.

\noindent\textbf{Dual Orthogonality Regularization.}
Naive sharing can cause future states to collapse toward the present state or become redundant
across horizons.
To prevent this, we regularize the residuals in Eq.~\eqref{eq:residual}.
After $\ell_2$ normalization,
\begin{equation}
\tilde{\mathbf{Z}}_{\mathrm{det},bi}
=
\frac{\bar{\mathbf{Z}}_{\mathrm{det},bi}}
{\|\bar{\mathbf{Z}}_{\mathrm{det},bi}\|_2 + \epsilon},
\qquad
\tilde{\mathbf{R}}_{bi}^{(h)}
=
\frac{\mathbf{R}_{bi}^{(h)}}
{\|\mathbf{R}_{bi}^{(h)}\|_2 + \epsilon},
\label{eq:orth_norm}
\end{equation}
where $\delta_b^{(h)}\in\{0,1\}$ indicates whether sample $b$ has valid supervision at horizon
$h$.

Task orthogonality separates present grounding from future change:
\begin{equation}
\mathcal{L}_{\mathrm{t\mbox{-}orth}}
=
\frac{
\sum_{b,h,i}
\delta_b^{(h)}
\Big(
\tilde{\mathbf{Z}}_{\mathrm{det},bi}^{\top}
\tilde{\mathbf{R}}_{bi}^{(h)}
\Big)^2
}{
\sum_{b,h,i}\delta_b^{(h)} + \epsilon
},
\label{eq:loss_torth}
\end{equation}
while horizon orthogonality separates future directions across horizons:
\begin{equation}
\mathcal{L}_{\mathrm{h\mbox{-}orth}}
=
\frac{1}{|\mathcal{P}_{\mathrm{valid}}|+\epsilon}
\sum_{(h_a,h_b)\in\mathcal{P}_{\mathrm{valid}}}
\frac{
\sum_{b,i}
\delta_b^{(h_a)}\delta_b^{(h_b)}
\Big(
(\tilde{\mathbf{R}}_{bi}^{(h_a)})^\top
\tilde{\mathbf{R}}_{bi}^{(h_b)}
\Big)^2
}{
\sum_{b,i}\delta_b^{(h_a)}\delta_b^{(h_b)}+\epsilon
},
\label{eq:loss_horth}
\end{equation}
where $\mathcal{P}_{\mathrm{valid}} \subseteq
\{(h_a,h_b)\mid h_a<h_b,\,h_a,h_b\in\mathcal{H}\}$ contains horizon pairs with at least one
valid sample; otherwise $\mathcal{L}_{\mathrm{h\mbox{-}orth}}=0$.

\noindent\textbf{Final Objective.}
Because reliable anticipation depends on stable pair grounding, we gradually ramp up all
anticipation-related terms with
\begin{equation}
\alpha(e)
=
\alpha_0
+
(1-\alpha_0)\,
\min\!\left(
\frac{e}{E_{\mathrm{warm}}-1},
\,1
\right),
\label{eq:rampup}
\end{equation}
and optimize
\begin{equation}
\mathcal{L}
=
\mathcal{L}_{\mathrm{det}}
+
\alpha(e)
\Big(
\mathcal{L}_{\mathrm{ant}}
+
\lambda_{\mathrm{t\mbox{-}orth}}
\mathcal{L}_{\mathrm{t\mbox{-}orth}}
+
\lambda_{\mathrm{h\mbox{-}orth}}
\mathcal{L}_{\mathrm{h\mbox{-}orth}}
\Big).
\label{eq:loss_total}
\end{equation}
Following DETR-style training~\cite{carion2020detr}, auxiliary supervision is applied to
intermediate decoder outputs using the same loss definitions.

\section{The DETAnt-HOI Benchmark}
\label{sec:benchmark}

\begin{figure}[t]
  \centering
  \includegraphics[width=\columnwidth]{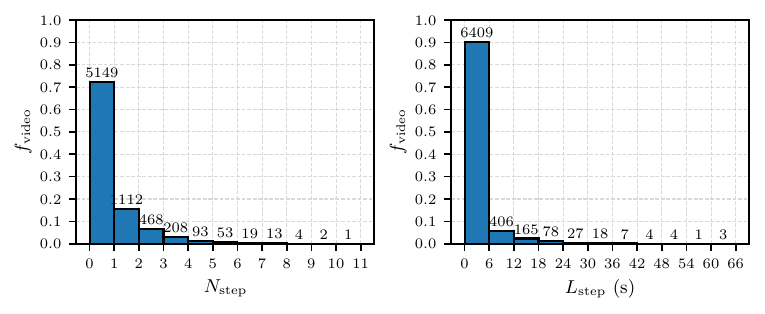}
  \vskip-3ex
  \caption{
    \textbf{Temporal non-continuity in the VidHOI evaluation protocol.}
    We analyze temporal gaps between annotated keyframes in VidHOI from two perspectives:
    (a) the number of temporal discontinuities per video ($N_{\text{step}}$), and
    (b) the duration of these gaps in seconds ($L_{\text{step}}$).
    A substantial portion of videos exhibits frequent and long temporal gaps,
    indicating that nominal ``future'' labels may not correspond to temporally coherent continuations of observed interactions.
  }
  \Description{
    The figure presents a statistical analysis of temporal discontinuities in the VidHOI evaluation protocol.
    The left histogram shows the distribution of the number of time gaps per video,
    while the right histogram shows the distribution of gap durations in seconds.
    Most videos contain at least one temporal discontinuity, and a non-negligible portion exhibits multiple gaps with large temporal spans.
    These gaps arise from the keyframe-based annotation scheme, where intermediate non-interactive frames are omitted, breaking temporal continuity.
    As a result, annotated ``future'' frames may be separated from the observed clip by irregular and sometimes long intervals,
    making them unreliable as true future supervision signals.
    This observation motivates a temporally corrected evaluation protocol that preserves instance continuity
    and ensures that anticipation is evaluated on genuinely future interaction dynamics rather than annotation artifacts.
  }
  \vskip-3ex
  \label{fig:dataset}
\end{figure}

Existing video HOI benchmarks do not provide a fully anticipation-faithful evaluation setting. In both VidHOI dataset~\cite{chiou2021vidhoi} and Action Genome dataset~\cite{ji2020action}, clips are commonly constructed from sparsely annotated keyframes, where non-interactive frames are discarded by prior evaluation protocols~\cite{chiou2021vidhoi, ni2023gaze}.
Under such protocols, the nominal ``future'' label is not always a temporally coherent continuation of the observed interaction.

Taking VidHOI as an example, Fig.~\ref{fig:dataset} reveals this temporal discontinuity: nearly $30\%$ of videos contain time gaps between consecutive keyframes.
These gaps can occur frequently (up to 11 times per video) and can be large (up to 66 seconds).
As a result, reported anticipation performance may partly reflect annotation sparsity and clip-construction artifacts rather than genuine future reasoning over continuous pair dynamics.

To address this issue, we establish \textbf{DETAnt-HOI}, a temporally corrected benchmark protocol for unified HOI detection and anticipation built on VidHOI and Action Genome.
Rather than modifying dataset splits or label spaces, DETAnt-HOI corrects the \emph{evaluation protocol}: it preserves instance continuity, avoids clips dominated by frames without interaction supervision, and enforces a consistent pair-level detection-and-anticipation setting across datasets.
Therefore, DETAnt-HOI should be understood not as a new label space, but as a corrected evaluation setting for studying whether anticipation is learned from continuous interaction dynamics rather than annotation-induced shortcuts.

\subsection{Benchmark Construction}
\label{sec:benchmark:construction}

We retain the original train/validation/test splits of both datasets and modify only the clip-construction procedure.
Each sample consists of an observation window of 6 \emph{keyframes}, with HOI detection defined on the last observed frame.
Future anticipation is evaluated at horizons
\(
\mathcal{H}=\{1,3,5,7\}.
\)

On VidHOI, these horizons correspond to temporal offsets in seconds.
On Action Genome, they correspond to approximately aligned future keyframe offsets under the adopted keyframe alignment protocol.

A valid clip must satisfy two conditions.
First, it must preserve temporal continuity for instance-level pair tracking.
Second, intermediate frames may contain no HOI interactions, but the last observed frame must contain HOI supervision, as it serves as the detection target.
Future supervision may be unavailable at certain horizons due to missing keyframes; such horizons are masked during both training and evaluation.

\subsection{Temporal Continuity Correction}
\label{sec:benchmark:continuity}

Rather than constructing clips solely from interaction-labeled key-frames, we explicitly handle non-interactive intervals.

If the number of consecutive non-interactive frames is smaller than the observation window, we retain these frames to avoid temporal gaps.
Their bounding-box annotations preserve instance continuity and improve subject--object tracking.

If the number of consecutive non-interactive frames exceeds the observation window, we remove these frames to ensure clip validity.
We additionally split the video at this interval and treat the resulting segments as independent sequences.
This is justified because interaction correlation becomes weak over long temporal gaps.

This procedure yields temporally continuous clips suitable for both detection and anticipation.

\noindent\textbf{Annotation Supplement for VidHOI.}
In VidHOI, non-interactive frames are removed in the released keyframe stream.
To recover instance continuity, we supplement \emph{frame-level instance annotations} for short non-interaction gaps as described above.
These annotations are obtained from the VidOR~\cite{shang2019VidOR} dataset, the original source of VidHOI annotations, ensuring consistency.

This supplementation is meaningful because non-interactive frames often still contain visible humans and objects, and the corresponding subject--object pairs may remain temporally continuous even without interaction labels.
Ignoring such frames breaks pair trajectories and makes clip construction unsuitable for faithful anticipation evaluation.
Statistics of the supplemented frames are reported in Table~\ref{tab:benchmark_stats}.

\subsection{Unified Protocol}
\label{sec:benchmark:protocol}

To enable a unified model and evaluation pipeline, we convert annotations into a frame-level HOI-A format~\cite{liao2020ppdm} and maintain separate metadata for temporal alignment.

Future pair alignment is established via IoU-based matching with a threshold of 0.5, following standard HOI evaluation protocols.
This results in an evaluation setting where detection and anticipation are measured on temporally consistent, pair-aligned interaction instances rather than raw sparse keyframe clips.

\subsection{Benchmark Statistics}
\label{sec:benchmark:stats}

Table~\ref{tab:benchmark_stats} summarizes DETAnt-HOI.
Compared with the original clip construction, the corrected protocol yields more temporally valid train/test clips by repairing short discontinuities and removing long inactive spans.

Beyond clip counts, the correction significantly reshapes the frame distribution.
On VidHOI, the number of frames increases (e.g., 193k $\rightarrow$ 199k for training), as short non-interaction gaps are filled.
In contrast, on Action Genome, the number of frames decreases (e.g., 218k $\rightarrow$ 192k), since long discontinuous segments are removed.

This asymmetric effect reflects two complementary operations:
recovering short-term continuity and eliminating long-range temporal breaks.
As a result, DETAnt-HOI provides a more reliable evaluation setting in which future labels correspond to temporally coherent interaction evolution rather than annotation artifacts.
Supplementary annotations are required only for VidHOI, while Action Genome requires only protocol-level correction.

\begin{table}[t]
  \centering
  \caption{
    \textbf{DETAnt-HOI Statistics.}
    ``Original'' and ``DETAnt-HOI'' denote clip counts before and after temporal continuity correction.
    For ``Valid pairs'' and ``Supplementary frames added'', values are reported as \textit{train/test}.
  }
  \vskip-2ex
  \label{tab:benchmark_stats}
  \small
  \setlength{\tabcolsep}{3.6pt}
  \renewcommand{\arraystretch}{1.08}
  \resizebox{\columnwidth}{!}{%
  \begin{tabular}{lcccc}
    \toprule
    & \multicolumn{2}{c}{\textbf{VidHOI}} & \multicolumn{2}{c}{\textbf{Action Genome}} \\
    \cmidrule(lr){2-3}\cmidrule(lr){4-5}
    & Original & DETAnt-HOI & Original & DETAnt-HOI \\
    \midrule
    Train clips   & 162952 & 165140 & 139453 & 145470 \\
    Test clips    & 19299  & 19542  & 48163  & 52748 \\
    Valid pairs @ $h{=}1$ & \multicolumn{2}{c}{50386 / 5564} & \multicolumn{2}{c}{22212 / 6936} \\
    Valid pairs @ $h{=}3$ & \multicolumn{2}{c}{46658 / 5160} & \multicolumn{2}{c}{20570 / 6677} \\
    Valid pairs @ $h{=}5$ & \multicolumn{2}{c}{43039 / 4711} & \multicolumn{2}{c}{18799 / 6332} \\
    Valid pairs @ $h{=}7$ & \multicolumn{2}{c}{40026 / 4310} & \multicolumn{2}{c}{17383 / 6051} \\
    Supplementary frames added & \multicolumn{2}{c}{4655 / 548} & \multicolumn{2}{c}{0 / 0} \\
    \bottomrule
  \end{tabular}%
  }
  \vskip-2ex
\end{table}
\section{Experiments}
\label{sec:experiment}
\begin{table*}[t!]
\centering
\caption{
\textbf{Comparison with prior video HOI methods on DETAnt-HOI.}
We report present-time detection mAP on the Full, Rare, and Non-rare splits, and future anticipation mAP at horizons \(h \in \mathcal{H}\), on both the VidHOI and Action Genome components.
}
\vskip-2ex
\label{tab:map}
\small
\setlength{\tabcolsep}{3.1pt}
\renewcommand{\arraystretch}{1.08}
\resizebox{\textwidth}{!}{%
\begin{tabular}{lllccccccc|ccccccc}
\toprule
\multirow{3}{*}{\textbf{Model}} &
\multirow{3}{*}{\textbf{Paradigm}} &
\multirow{3}{*}{\textbf{Detector}} &
\multicolumn{7}{c|}{\textbf{VidHOI Component}} &
\multicolumn{7}{c}{\textbf{Action Genome Component}} \\
\cmidrule(lr){4-10} \cmidrule(lr){11-17}
& & &
\multicolumn{3}{c}{\textbf{Det. mAP} $\uparrow$} &
\multicolumn{4}{c|}{\textbf{Ant. mAP} $\uparrow$} &
\multicolumn{3}{c}{\textbf{Det. mAP} $\uparrow$} &
\multicolumn{4}{c}{\textbf{Ant. mAP} $\uparrow$} \\
\cmidrule(lr){4-6} \cmidrule(lr){7-10}
\cmidrule(lr){11-13} \cmidrule(lr){14-17}
& & &
\textbf{Full} & \textbf{Rare} & \textbf{Non-rare} &
\textbf{$h{=}1$} & \textbf{$h{=}3$} & \textbf{$h{=}5$} & \textbf{$h{=}7$} &
\textbf{Full} & \textbf{Rare} & \textbf{Non-rare} &
\textbf{$h{=}1$} & \textbf{$h{=}3$} & \textbf{$h{=}5$} & \textbf{$h{=}7$} \\
\midrule
ST-HOI~\cite{chiou2021vidhoi} & Two-Stage & Faster R-CNN~\cite{ren2016faster}
& 3.10 & 2.10 & 5.90 & -- & -- & -- & --
& -- & -- & -- & -- & -- & -- & -- \\

STTran~\cite{cong2021sttran} & Two-Stage & Faster R-CNN~\cite{ren2016faster}
& 7.61 & 3.33 & 13.18 & 8.80 & 8.32 & 8.67 & 8.75
& 6.11 & 0.20 & 7.30 & 6.07 & 5.52 & 5.24 & 4.92 \\

Gaze-Tran~\cite{ni2023gaze} & Two-Stage & YOLOv5~\cite{khanam2024yolov5}
& 10.40 & 5.46 & 16.83 & 11.30 & 10.65 & 10.19 & 10.14
& 6.99 & 0.50 & 9.86 & 8.02 & 7.17 & 6.70 & 6.47 \\

\best{HOI-DA} & \best{One-Stage} & \best{--}
& \best{16.27} & \best{12.21} & \best{22.35}
& \best{16.40} & \best{16.02} & \best{16.63} & \best{18.73}
& \best{9.70} & \best{1.88} & \best{13.32}
& \best{9.22} & \best{8.48} & \best{8.08} & \best{7.60} \\
\bottomrule
\end{tabular}%
}
\vskip-2ex
\end{table*}

\begin{table*}[t!]
\centering
\caption{
\textbf{Recall-based evaluation on the VidHOI component of DETAnt-HOI.}
We report Recall@10, Recall@20, and Recall@50 for present-time prediction (\(h{=}0\)) and future anticipation at horizons \(h \in \mathcal{H}\).
}
\vskip-2ex
\label{tab:vidhoi}
\small
\setlength{\tabcolsep}{3.8pt}
\renewcommand{\arraystretch}{1.08}
\resizebox{\textwidth}{!}{%
\begin{tabular}{llccccc|ccccc|ccccc}
\toprule
\multirow{2}{*}{\textbf{Model}} & \multirow{2}{*}{\textbf{Paradigm}} &
\multicolumn{5}{c|}{\textbf{Recall@10} $\uparrow$} &
\multicolumn{5}{c|}{\textbf{Recall@20} $\uparrow$} &
\multicolumn{5}{c}{\textbf{Recall@50} $\uparrow$} \\
\cmidrule(lr){3-7}\cmidrule(lr){8-12}\cmidrule(lr){13-17}
& &
\textbf{$h{=}0$} & \textbf{$h{=}1$} & \textbf{$h{=}3$} & \textbf{$h{=}5$} & \textbf{$h{=}7$} &
\textbf{$h{=}0$} & \textbf{$h{=}1$} & \textbf{$h{=}3$} & \textbf{$h{=}5$} & \textbf{$h{=}7$} &
\textbf{$h{=}0$} & \textbf{$h{=}1$} & \textbf{$h{=}3$} & \textbf{$h{=}5$} & \textbf{$h{=}7$} \\
\midrule
STTran~\cite{cong2021sttran} & Two-Stage
& 39.56 & 38.96 & 38.18 & 37.35 & 37.13
& 40.06 & 38.76 & 38.22 & 39.38 & 40.31
& 42.75 & 41.90 & 42.69 & 42.13 & 47.02 \\

Gaze-Tran~\cite{ni2023gaze} & Two-Stage
& 46.20 & 47.14 & 46.98 & 48.08 & 47.66
& 48.94 & 50.00 & 50.03 & 50.90 & 50.66
& 49.62 & 50.78 & 50.90 & 51.65 & 51.41 \\

\best{HOI-DA} & \best{One-Stage}
& \best{57.54} & \best{58.08} & \best{59.01} & \best{59.60} & \best{59.92}
& \best{61.50} & \best{62.08} & \best{62.98} & \best{63.56} & \best{64.03}
& \best{63.29} & \best{63.97} & \best{64.73} & \best{65.25} & \best{65.68} \\
\bottomrule
\end{tabular}%
}
\vskip-2ex
\end{table*}

\begin{table*}[t!]
\centering
\caption{
\textbf{Recall-based evaluation on the Action Genome component of DETAnt-HOI.}
We report Recall@10, Recall@20, and Recall@50 for present-time prediction (\(h{=}0\)) and future anticipation at horizons \(h \in \mathcal{H}\).
}
\vskip-2ex
\label{tab:ag}
\small
\setlength{\tabcolsep}{3.8pt}
\renewcommand{\arraystretch}{1.08}
\resizebox{\textwidth}{!}{%
\begin{tabular}{llccccc|ccccc|ccccc}
\toprule
\multirow{2}{*}{\textbf{Model}} & \multirow{2}{*}{\textbf{Paradigm}} &
\multicolumn{5}{c|}{\textbf{Recall@10} $\uparrow$} &
\multicolumn{5}{c|}{\textbf{Recall@20} $\uparrow$} &
\multicolumn{5}{c}{\textbf{Recall@50} $\uparrow$} \\
\cmidrule(lr){3-7}\cmidrule(lr){8-12}\cmidrule(lr){13-17}
& &
\textbf{$h{=}0$} & \textbf{$h{=}1$} & \textbf{$h{=}3$} & \textbf{$h{=}5$} & \textbf{$h{=}7$} &
\textbf{$h{=}0$} & \textbf{$h{=}1$} & \textbf{$h{=}3$} & \textbf{$h{=}5$} & \textbf{$h{=}7$} &
\textbf{$h{=}0$} & \textbf{$h{=}1$} & \textbf{$h{=}3$} & \textbf{$h{=}5$} & \textbf{$h{=}7$} \\
\midrule
STTran~\cite{cong2021sttran} & Two-Stage
& 19.86 & 20.42 & 18.24 & 17.78 & 17.11
& 28.76 & 29.23 & 28.02 & 27.14 & 26.89
& 33.53 & 34.94 & 33.51 & 32.64 & 32.04 \\

Gaze-Tran~\cite{ni2023gaze} & Two-Stage
& 21.24 & 21.16 & 20.13 & 19.55 & 19.02
& 30.88 & 31.15 & 30.56 & 29.64 & 28.92
& 36.74 & 37.52 & 36.35 & 35.35 & 34.53 \\

\best{HOI-DA} & \best{One-Stage}
& \best{28.89} & \best{29.06} & \best{28.29} & \best{27.68} & \best{27.35}
& \best{34.70} & \best{35.17} & \best{34.53} & \best{34.14} & \best{33.85}
& \best{39.99} & \best{40.98} & \best{40.58} & \best{40.38} & \best{40.08} \\
\bottomrule
\end{tabular}%
}
\vskip-2ex
\end{table*}

\begin{figure*}[ht]
  \centering
  \includegraphics[width=\textwidth]{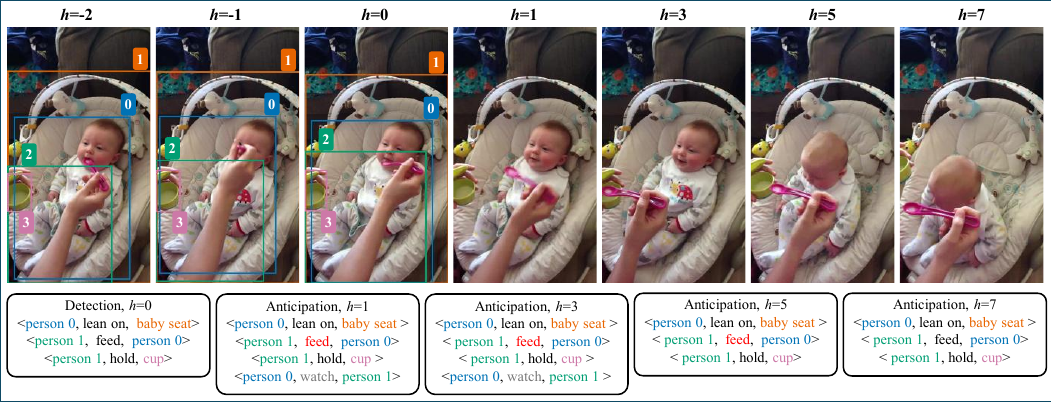}
  \vskip-2ex
  \caption{Qualitative results of unified HOI detection and multi-horizon anticipation on VidHOI. Given an observed video clip (6 frames, 3 shown), the model jointly predicts present interactions ($h=0$) and future HOIs at horizons $h \in \mathcal{H}$. Each row corresponds to a persistent human-object pair hypothesis maintained across time.  Detection results are reported at $h=0$ for consistency. Verb colors: black—true positives, red—false positives, gray—false negatives.
  }
  \Description{\Description{The figure visualizes joint human-object interaction (HOI) detection and multi-horizon anticipation on a video sequence. 
  The top row shows sampled frames across time, including both observed frames ($h \leq 0$) and future horizons ($h > 0$). 
  The bottom panels present predicted HOI triplets for each horizon. 
  Each interaction is organized by persistent human--object pair hypotheses, where the same pair identity is maintained across all time steps. 
  Present-time detection is performed at $h=0$, and future interactions are predicted for increasing horizons based on the observed clip. 
  The visualization highlights that dominant interactions (e.g., feeding and holding) remain stable across time, while transient interactions (e.g., watching) emerge at short horizons and disappear at longer horizons. 
  This demonstrates that the model performs history-conditioned forecasting rather than naive last-frame extrapolation. 
  Overall, the figure provides qualitative evidence that the proposed unified formulation preserves pair consistency and produces temporally coherent interaction predictions across multiple future horizons.
}}
  \vskip-2ex
  \label{fig:visual}
  \vskip-1ex
\end{figure*}

\subsection{Evaluation Metrics}
\label{subsec:data}

We evaluate HOI-DA on DETAnt-HOI, our temporally corrected benchmark built from VidHOI and Action Genome.
Following standard HOI evaluation, we use mean Average Precision (mAP) as the primary metric for both present-time detection and future anticipation.
A predicted HOI triplet is counted as correct if the predicted human box and object box each overlap the matched ground-truth boxes with IoU \(>0.5\), the object category is correct, and the predicate label is correct~\cite{chiou2021vidhoi,ni2023gaze}.
For the VidHOI component, we additionally report Full, Rare, and Non-rare mAP following the official protocol~\cite{chiou2021vidhoi}.
To assess whether the correct future HOI remains highly ranked as temporal uncertainty increases, we also report Recall@k for anticipation.
On VidHOI, this follows the person-wise top-$k$ recall protocol of Gaze-Tran~\cite{ni2023gaze}; on Action Genome, the same computation is equivalent to frame-wise Recall@k under the adopted single-subject HOI setting.
In all recall tables, \(h=0\) denotes prediction at the observation boundary and \(h>0\) denotes future anticipation.
Together, mAP and Recall@k measure not only present-time grounding, but also how well pair-level interaction hypotheses remain predictive across future horizons.

\subsection{Baselines}
\label{subsec:baselines}

We compare HOI-DA with representative prior video HOI methods covering the dominant design paradigms in the literature.
\textbf{ST-HOI}~\cite{chiou2021vidhoi} is an early spatio-temporal baseline introduced with VidHOI, based on external detection followed by temporal interaction reasoning.
\textbf{STTran}~\cite{cong2021sttran} represents spatio-temporal Transformer baselines that model inter-frame relational context on top of off-the-shelf detectors.
\textbf{Gaze-Tran}~\cite{ni2023gaze} extends this line with gaze-following cues and is one of the few published methods reporting both HOI detection and multi-horizon anticipation on VidHOI.

These baselines are all built around two-stage reasoning over externally formed human--object candidates.
In contrast, HOI-DA is a one-stage pair-centric framework that jointly optimizes observation-boundary detection and future anticipation within a shared representation space.
This comparison therefore tests not only accuracy, but also whether anticipation is better treated as a downstream prediction task or as a structural constraint on pair-level video representation learning.

\subsection{Implementation Details}
\label{subsec:implementation}

Unless otherwise specified, we use the same architecture and training protocol on both components of DETAnt-HOI.
HOI-DA adopts a ResNet-50~\cite{he2016deep} backbone and a DETR-style encoder--decoder Transformer~\cite{carion2020detr} with hidden dimension 384 and 180 learned queries.
We use a 6-frame observation window and predict future verbs at horizons \(\mathcal{H}=\{1,3,5,7\}\), matching the benchmark protocol.
We train for 50 epochs with AdamW~\cite{loshchilov2017decoupled}, using learning rates of \(10^{-4}\) for the Transformer and prediction heads and \(10^{-5}\) for the backbone and RoBERTa-based text encoder.
The batch size is 4 on 4 Blackwell GPUs, and the learning rate is decayed by \(0.1\) after epoch 30.
Unless otherwise stated, the text encoder is fine-tuned and uses the \texttt{vidhoi\_natural} prompt style.
For anticipation, the overall loss coefficient is \(0.8\), the horizon decay factor is \(0.7\), both orthogonality coefficients are \(0.05\), and the anticipation ramp increases linearly from \(0.25\) to \(1.0\) over the first 8 epochs.
This configuration emphasizes near-term prediction while preserving longer-horizon supervision and stabilizing joint optimization of detection and anticipation.

\subsection{Comparison to State of the Art}
\label{subsec:comparison}

Tables~\ref{tab:map}--\ref{tab:ag} compare HOI-DA with prior video HOI methods on DETAnt-HOI.
HOI-DA achieves the best results on all reported detection and anticipation metrics across both benchmark components.
More importantly, its advantage grows as the forecasting horizon increases, which is the central empirical pattern of this paper: the proposed formulation improves not only present-time grounding, but also the temporal persistence of pair-level interaction representations.
On the VidHOI component, HOI-DA surpasses the strongest prior baseline, Gaze-Tran, by \(+5.10\) mAP at \(h=1\) and \(+8.59\) mAP at \(h=7\).
This widening margin indicates that prior two-stage pipelines become increasingly brittle as temporal uncertainty grows, whereas HOI-DA remains substantially more stable.
The same pattern is visible in Recall@k (Tables~\ref{tab:vidhoi} and~\ref{tab:ag}), showing that the gain is not merely due to score calibration, but to stronger ranking of plausible future HOIs over the same pair hypotheses.
The detection results are equally consistent.
On VidHOI, HOI-DA improves over Gaze-Tran by \(+5.87\), \(+6.75\), and \(+5.52\) mAP on the Full, Rare, and Non-rare splits, respectively.
On Action Genome, the absolute margins are smaller but remain positive across all reported metrics for both detection and anticipation.
Taken together, these results support the main claim of this work: anticipation is most effective when it is not treated as a downstream add-on, but used to shape the pair representation itself.
In this sense, HOI-DA is not simply a stronger predictor, but a stronger formulation of joint video HOI detection and anticipation.

\subsection{Ablation Study}
\label{subsec:ablation}

Table~\ref{tab:ablation} shows that the gain of HOI-DA does not come from naive task fusion.
A unified one-decoder design performs worse than even the separate detection/anticipation baseline, indicating that unrestricted sharing introduces interference rather than useful transfer.
The full model substantially outperforms both, showing that joint modeling is beneficial only when present grounding and future forecasting are coupled in a structured manner.

The Temporal Summary Module is most important for long-range anticipation.
Removing it degrades all horizons, with the gap increasing toward \(h=7\), which confirms that future verb prediction should be conditioned on the observed interaction history rather than on the boundary frame alone.

The text-related ablations reveal complementary roles of semantic structure.
Removing only decoder-side text guidance causes a modest drop, whereas removing the full language-guided semantic branch leads to a much larger degradation in both detection and anticipation.
This indicates that language primarily helps by regularizing the shared interaction space, while decoder-side semantic guidance provides an additional but smaller gain.

Dual orthogonality regularization mainly affects long-horizon forecasting.
Its removal has limited impact on detection but increasingly harms anticipation as the horizon grows, consistent with its role in separating present grounding, future change, and horizon-specific dynamics.
Overall, the ablations show that HOI-DA works not because detection and anticipation are merely trained together, but because they are coupled through a representation explicitly structured for pair continuity over time.

\begin{table}[t]
\centering
\caption{
\textbf{Ablation study on the VidHOI of DETAnt-HOI.}
}
\vskip-2ex
\label{tab:ablation}
\small
\setlength{\tabcolsep}{3.6pt}
\renewcommand{\arraystretch}{1.08}
\resizebox{\linewidth}{!}{%
\begin{tabular}{lccc|cccc}
\toprule
\multirow{2}{*}{\textbf{Variant}} &
\multicolumn{3}{c|}{\textbf{Det. mAP} $\uparrow$} &
\multicolumn{4}{c}{\textbf{Ant. mAP} $\uparrow$} \\
\cmidrule(lr){2-4}\cmidrule(lr){5-8}
& \textbf{Full} & \textbf{Rare} & \textbf{Non-rare}
& \textbf{$h{=}1$} & \textbf{$h{=}3$} & \textbf{$h{=}5$} & \textbf{$h{=}7$} \\
\midrule
Separate Detection and Anticipation
& 14.29 & 10.83 & 19.00 & 14.09 & 13.28 & 13.36 & 14.20 \\

Unified One-Decoder
& 13.08 & 9.26 & 18.64 & 12.61 & 13.79 & 14.63 & 14.66 \\

w/o Temporal Summary Module
& 15.58 & 11.76 & 21.31 & 14.95 & 14.96 & 15.66 & 16.05 \\

w/o Decoder-Side Text Guidance
& 16.11 & 12.43 & 21.64 & 16.31 & 15.26 & 15.67 & 16.38 \\

w/o Language-Guided Semantic Branch
& 13.93 & 10.70 & 18.77 & 14.26 & 14.09 & 16.09 & 16.48 \\

w/o Dual Orthogonality Regularization
& 15.97 & 11.91 & 22.07 & 15.85 & 14.32 & 14.63 & 16.39 \\

\best{HOI-DA (full)}
& \best{16.27} & \best{12.21} & \best{22.35}
& \best{16.40} & \best{16.02} & \best{16.63} & \best{18.73} \\
\bottomrule
\end{tabular}%
}
\vskip-2ex
\end{table}

\subsection{Qualitative Results}
\label{subsec:visual}

Figure~\ref{fig:visual} illustrates joint HOI detection and multi-horizon anticipation on VidHOI. Unlike detect/track-then-forecast pipelines that operate on externally constructed human--object pairs, HOI-DA maintains persistent pair hypotheses and predicts both present and future interactions within a shared pair-centric representation.

\noindent\textbf{(1) Temporally persistent pair representation.}
As shown in Fig.~\ref{fig:visual}, HOI-DA preserves consistent human--object pair identity from the observation window to all future horizons. The caregiver--baby and caregiver--cup interactions remain anchored to the same pair slots throughout the sequence. This contrasts with two-stage pipelines, where pair construction depends on external detection and tracking, often leading to unstable associations under motion or occlusion. By treating pair identity as an internal representation rather than an external pre-processing step, HOI-DA enables temporally consistent reasoning over the same interaction instance.

\noindent\textbf{(2) History-conditioned future prediction.}
HOI-DA does not simply extrapolate from the last observed frame. In Fig.~\ref{fig:visual}, the additional \textit{watch} relation between person~0 and person~1 emerges at short-term horizons ($h=1$ and $h=3$) but disappears at longer horizons ($h=5$ and $h=7$), while the primary interactions remain stable. This behavior indicates that future predictions are conditioned on the full interaction history rather than copied from the boundary state. In contrast, baseline methods that rely on boundary-based extrapolation tend to repeat current interactions or fail to capture such transient relations.

\noindent\textbf{(3) Stability at long horizons.}
The qualitative results further highlight that HOI-DA maintains coherent predictions even at longer horizons. While two-stage methods typically degrade as temporal uncertainty increases, HOI-DA preserves plausible interaction dynamics across all horizons, consistent with the quantitative trend where performance gaps widen over time. This supports the view that modeling future interactions as residual transitions from present pair states leads to more robust long-term reasoning.

Overall, Fig.~\ref{fig:visual} provides direct visual evidence for our central claim: anticipation should not be treated as a downstream prediction task, but as a structural constraint on pair-level video representation. By preserving pair continuity and modeling future interactions as structured evolutions of current pair states, HOI-DA produces temporally consistent forecasts beyond the observation boundary.

\section{Conclusion}
\label{sec:conclusion}
We presented a unified view of video HOI detection and anticipation, arguing that future prediction should serve as a structural constraint on pair-level video representation learning rather than a downstream add-on. Based on this idea, we introduced HOI-DA, which models future HOIs as residual evolutions of present pair states, and DETAnt-HOI, which corrects temporal discontinuities in existing evaluation protocols. Experiments on VidHOI and Action Genome show consistent improvements in both detection and anticipation, especially at longer horizons, while ablations confirm that the gains come from structured coupling rather than naive task fusion. Overall, our results highlight the importance of both stronger formulations and more faithful evaluation for progress in video HOI anticipation.

\begin{acks}
This work was supported in part by the Deutsche Forschungsgemeinschaft (DFG, German Research Foundation) - SFB 1574 - 471687386. 
\end{acks}

\clearpage
\balance
\bibliographystyle{ACM-Reference-Format}
\bibliography{acmart}


\clearpage
\appendix
\section{Additional Details on the DETAnt-HOI Benchmark}
\label{sec:supp-benchmark}
 
As described in Section~4 of the main paper, DETAnt-HOI enforces temporal
continuity by preserving short non-interactive intervals and splitting videos
at long inactive gaps.
The concrete procedure differs between VidHOI and Action Genome due to
structural differences in their original annotations.
 
\noindent\textbf{VidHOI.}
In the released VidHOI benchmark, annotated frame indices are sparse:
the published annotation files omit short unlabeled intervals between
HOI-labeled key-frames, so the released key-frame stream contains no
bounding-box supervision for those gaps.
To recover instance continuity across such intervals, we supplement them
with frame-level bounding-box annotations sourced from
VidOR~\cite{shang2019VidOR}, the original dataset from which all VidHOI
videos and labels are derived, maintaining consistency with the original
VidOR annotations.
Even in the absence of an active HOI label, the relevant human and object
instances remain physically present in the scene, and their positional
continuity is essential for stable pair tracking across the observation
window.
Ignoring these gaps severs pair trajectories and renders clip construction
anticipation-unfaithful, as evidenced quantitatively in Table~1 of the main
paper.
 
\noindent\textbf{Action Genome.}
Unlike VidHOI, the released Action Genome annotations already include
frames annotated with negative or transitional relations (e.g.,
\textit{not contacting}, \textit{not looking at}), so no supplementary
external annotation is required.
Our correction operates at the protocol level: we identify long inactive
intervals (i.e., consecutive key-frame spans carrying no HOI supervision
that exceed the clip length), filter their frames from clip construction,
and split the underlying video into two independent segments at such
boundaries.
This is reasonable because pair-level HOI correlations become negligible
after a sufficiently long inactive interval.
The correction does not modify the original annotation files; it is
implemented entirely through a video-level metadata file that records
segment boundaries and enumerates the key-frames belonging to each
reconstructed clip, leaving source annotations intact.

\section{Comparison with Image-Based HOI Methods}
\label{sec:supp-image}
 
Because publicly available video HOI models with open-source code are scarce,
we supplement the main comparison with two representative open-source
end-to-end image-based HOI detectors: QPIC~\cite{tamura2021qpic}, a purely
visual one-stage model, and RLIPv2-ParSeDA~\cite{yuan2023rlipv2}, a
one-stage vision--language pre-training approach.
Note that image-based HOI detection encompasses both one-stage and two-stage
methods; we select these two one-stage models specifically to avoid the
confound of external detector quality and to enable a clean comparison of
temporal context versus appearance-only reasoning.
For a fair evaluation, each video key-frame is treated as an independent
image input, and we ensure that the total number of training images seen per
epoch is approximately equal across image-based and video-based methods.
HOI-DA is evaluated here with a ResNet-50~\cite{he2016deep} backbone to
control for feature capacity; the effect of backbone choice is studied
separately in Section~\ref{sec:supp-backbone}.
 
\noindent\textbf{Detection mAP (Table~\ref{tab:supp-image-map}).}
Image-based methods fall consistently below HOI-DA on Full, Rare, and
Non-rare mAP, demonstrating that temporal context provides a meaningful
signal beyond single-frame appearance even for the present-time detection
task.
Anticipation columns are not applicable for image-based methods by design,
since they operate on a single frame without access to the observed
interaction history.
 
\noindent\textbf{Recall@$k$ (Table~\ref{tab:supp-image-recall}).}
The results reveal an asymmetry between mAP and Recall@$k$ that is
informative in itself.
HOI-DA leads on mAP and on Recall@10 across all conditions.
However, at higher $k$ for present-time detection ($h{=}0$), image-based
methods are competitive and can exceed HOI-DA: RLIPv2-ParSeDA achieves
Recall@20 of 61.92 versus HOI-DA's 61.50, and both QPIC (65.06) and
RLIPv2-ParSeDA (68.88) surpass HOI-DA (63.29) at Recall@50.
This pattern reflects a structural property of the two metrics rather than a
failure of temporal modeling.
mAP penalizes fine-grained predicate confusion (e.g., mistaking
\textit{hold} for \textit{touch}), which is precisely where temporal
context helps most, since many near-synonymous predicates are only
disambiguated by motion cues across frames.
Recall@$k$ at large $k$, by contrast, only requires the correct interaction
to appear somewhere in the top-$k$ list. Even a model without temporal
context can cover enough plausible interactions to include the ground truth
when $k$ is large, which explains why image-based methods remain competitive
on high-$k$ present-time recall despite their lower mAP.
As the anticipation horizon increases ($h{>}0$), image-based methods produce
no output by design and HOI-DA leads across all $k$.


\begin{table}[h]
\centering
\caption{%
  \textbf{Detection mAP comparison with image-based HOI methods on the
  VidHOI component of DETAnt-HOI.}
  Image-based methods are evaluated by treating each key-frame as an
  independent image input.
}
\vskip-2ex
\label{tab:supp-image-map}
\small
\setlength{\tabcolsep}{5pt}
\renewcommand{\arraystretch}{1.10}
\resizebox{\linewidth}{!}{%
\begin{tabular}{llc ccc}
\toprule
\multirow{2}{*}{\textbf{Model}}
  & \multirow{2}{*}{\textbf{Paradigm}}
  & \multirow{2}{*}{\textbf{Modality}}
  & \multicolumn{3}{c}{\textbf{Det.\ mAP} $\uparrow$} \\
\cmidrule(lr){4-6}
  & & &
  \textbf{Full} & \textbf{Rare} & \textbf{Non-rare} \\
\midrule
QPIC~\cite{tamura2021qpic}
  & One-Stage & Image
  & 10.04 & 4.76  & 16.92 \\
RLIPv2-ParSeDA~\cite{yuan2023rlipv2}
  & One-Stage & Image + Text
  & 13.52 & 7.87  & 20.86 \\
\midrule
\best{HOI-DA (ResNet-50)}
  & \best{One-Stage} & \best{Video + Text}
  & \best{16.27} & \best{12.21} & \best{22.35}\\
\bottomrule
\end{tabular}%
}
\vskip-2ex
\end{table}

\begin{table}[h]
\centering
\caption{%
  Present-time ($h{=}0$) Recall@$k$ comparison with image-based
  HOI methods on the VidHOI component of DETAnt-HOI.}
  \vskip-2ex
\label{tab:supp-image-recall}
\small
\setlength{\tabcolsep}{8pt}
\renewcommand{\arraystretch}{1.10}
\resizebox{\linewidth}{!}{%
\begin{tabular}{llc ccc}
\toprule
\textbf{Model}
  & \textbf{Paradigm}
  & \textbf{Modality}
  & \textbf{Recall@10} $\uparrow$
  & \textbf{Recall@20} $\uparrow$
  & \textbf{Recall@50} $\uparrow$ \\
\midrule
QPIC~\cite{tamura2021qpic}
  & One-Stage & Image
  & 50.94 & 58.90 & 65.06 \\
RLIPv2-ParSeDA~\cite{yuan2023rlipv2}
  & One-Stage & Image + Text
  & 53.53 & \best{61.92} & \best{68.88} \\
\midrule
HOI-DA (ResNet-50)
  & One-Stage & Video + Text
  & \best{57.54} & 61.50 & 63.29 \\
\bottomrule
\end{tabular}}
\vskip-2ex
\end{table}

\section{Backbone Sensitivity of HOI-DA}
\label{sec:supp-backbone}

To isolate the effect of visual feature capacity from the comparison with
image-based methods, Table~\ref{tab:supp-backbone} evaluates HOI-DA under
two backbone configurations: ResNet-50~\cite{he2016deep} and
Swin-T~\cite{liu2021swin}.
Both backbones are trained end-to-end within the HOI-DA framework under the
same hyperparameter settings described in Section~5.3 of the main paper.
 
Despite its higher representational capacity in image recognition, Swin-T
underperforms ResNet-50 on both detection and anticipation under the current
integration.
We attribute this to a limitation of the current implementation: only the
final feature map of Swin-T is forwarded to the spatio-temporal encoder,
forgoing the multi-scale feature hierarchy that is central to the Swin
design.
This suggests that the HOI-DA framework does not yet benefit from increased
backbone capacity in its current form, and that a more careful integration
of multi-scale Swin features, or a dedicated feature pyramid, would be
needed to realize the potential of stronger backbones.
Incorporating such extensions remains a natural direction for future work.

\begin{table*}[t!]
\centering
\caption{%
  \textbf{Backbone sensitivity of HOI-DA on DETAnt-HOI.}
  Both variants use identical hyperparameters.
}
\vskip-2ex
\label{tab:supp-backbone}
\small
\setlength{\tabcolsep}{5pt}
\renewcommand{\arraystretch}{1.10}
\resizebox{\textwidth}{!}{%
\begin{tabular}{lc ccc cccc ccc cccc}
\toprule
\multirow{3}{*}{\textbf{Backbone}}
  & \multirow{3}{*}{\textbf{Modality}}
  & \multicolumn{7}{c}{\textbf{VidHOI Component}}
  & \multicolumn{7}{c}{\textbf{Action Genome Component}} \\
\cmidrule(lr){3-9}\cmidrule(lr){10-16}
  & &
  \multicolumn{3}{c}{\textbf{Det.\ mAP} $\uparrow$} &
  \multicolumn{4}{c}{\textbf{Ant.\ mAP} $\uparrow$} &
  \multicolumn{3}{c}{\textbf{Det.\ mAP} $\uparrow$} &
  \multicolumn{4}{c}{\textbf{Ant.\ mAP} $\uparrow$} \\
\cmidrule(lr){3-5}\cmidrule(lr){6-9}
\cmidrule(lr){10-12}\cmidrule(lr){13-16}
  & &
  \textbf{Full} & \textbf{Rare} & \textbf{Non-rare} &
  \textbf{$h{=}1$} & \textbf{$h{=}3$} & \textbf{$h{=}5$} & \textbf{$h{=}7$} &
  \textbf{Full} & \textbf{Rare} & \textbf{Non-rare} &
  \textbf{$h{=}1$} & \textbf{$h{=}3$} & \textbf{$h{=}5$} & \textbf{$h{=}7$} \\
\midrule
ResNet-50~\cite{he2016deep}
  & Video + Text
  & \best{16.27} & \best{12.21} & \best{22.35}
  & \best{16.40} & \best{16.02} & \best{16.63} & \best{18.73}
  & \best{9.70}  & \best{1.88}  & \best{13.32}
  & \best{9.22}  & \best{8.48}  & \best{8.08}  & \best{7.60} \\
Swin-T~\cite{liu2021swin}
  & Video + Text
  & 14.38 & 10.00 & 11.39 
  & 15.28 & 15.01 & 16.07 & 16.69
  & 9.50 & 1.79 & 11.62
  & 8.18  & 7.83 & 7.45 & 7.32 \\
\bottomrule
\end{tabular}%
}
\end{table*}

\begin{table*}[t!]
\centering
\caption{
\textbf{Recall-based evaluation on the VidHOI component of DETAnt-HOI.}
We report Recall@10, Recall@20, and Recall@50 for present-time prediction (\(h{=}0\)) and future anticipation at horizons \(h \in \mathcal{H}\).
}
\vskip-2ex
\label{tab:supp_recall_vidhoi}
\small
\setlength{\tabcolsep}{3.8pt}
\renewcommand{\arraystretch}{1.08}
\resizebox{\textwidth}{!}{%
\begin{tabular}{llccccc|ccccc|ccccc}
\toprule
\multirow{2}{*}{\textbf{Model}} & \multirow{2}{*}{\textbf{Data}} &
\multicolumn{5}{c|}{\textbf{Recall@10} $\uparrow$} &
\multicolumn{5}{c|}{\textbf{Recall@20} $\uparrow$} &
\multicolumn{5}{c}{\textbf{Recall@50} $\uparrow$} \\
\cmidrule(lr){3-7}\cmidrule(lr){8-12}\cmidrule(lr){13-17}
& &
\textbf{$h{=}0$} & \textbf{$h{=}1$} & \textbf{$h{=}3$} & \textbf{$h{=}5$} & \textbf{$h{=}7$} &
\textbf{$h{=}0$} & \textbf{$h{=}1$} & \textbf{$h{=}3$} & \textbf{$h{=}5$} & \textbf{$h{=}7$} &
\textbf{$h{=}0$} & \textbf{$h{=}1$} & \textbf{$h{=}3$} & \textbf{$h{=}5$} & \textbf{$h{=}7$} \\
\midrule

HOI-DA (ResNet-50~\cite{he2016deep}) & Video(+Text)
& \best{57.54} & \best{58.08} & \best{59.01} & \best{59.60} & \best{59.92}
& \best{61.50} & \best{62.08} & \best{62.98} & \best{63.56} & \best{64.03}
& \best{63.29} & \best{63.97} & \best{64.73} & \best{65.25} & \best{65.68} \\

HOI-DA (Swin-T~\cite{liu2021swin}) & Video(+Text)
& 54.35 & 54.94 & 49.58 & 56.28 & 56.43
& 58.32 & 58.99 & 59.82 & 60.33 & 60.45
& 59.98 & 60.53 & 61.26 & 61.78 & 61.92 \\
\bottomrule
\end{tabular}%
}
\end{table*}

\begin{table*}[t!]
\centering
\caption{
\textbf{Recall-based evaluation on the Action Genome component of DETAnt-HOI.}
We report Recall@10, Recall@20, and Recall@50 for present-time prediction (\(h{=}0\)) and future anticipation at horizons \(h \in \mathcal{H}\).
}
\vskip-2ex
\label{tab:supp_recall_ag}
\small
\setlength{\tabcolsep}{3.8pt}
\renewcommand{\arraystretch}{1.08}
\resizebox{\textwidth}{!}{%
\begin{tabular}{llccccc|ccccc|ccccc}
\toprule
\multirow{2}{*}{\textbf{Model}} & \multirow{2}{*}{\textbf{Data}} &
\multicolumn{5}{c|}{\textbf{Recall@10} $\uparrow$} &
\multicolumn{5}{c|}{\textbf{Recall@20} $\uparrow$} &
\multicolumn{5}{c}{\textbf{Recall@50} $\uparrow$} \\
\cmidrule(lr){3-7}\cmidrule(lr){8-12}\cmidrule(lr){13-17}
& &
\textbf{$h{=}0$} & \textbf{$h{=}1$} & \textbf{$h{=}3$} & \textbf{$h{=}5$} & \textbf{$h{=}7$} &
\textbf{$h{=}0$} & \textbf{$h{=}1$} & \textbf{$h{=}3$} & \textbf{$h{=}5$} & \textbf{$h{=}7$} &
\textbf{$h{=}0$} & \textbf{$h{=}1$} & \textbf{$h{=}3$} & \textbf{$h{=}5$} & \textbf{$h{=}7$} \\
\midrule

HOI-DA (ResNet-50~\cite{he2016deep}) & Video(+Text)
& \best{28.89} & \best{29.06} & \best{28.29} & \best{27.68} & \best{27.35}
& \best{34.70} & \best{35.17} & \best{34.53} & \best{34.14} & \best{33.85}
& \best{39.99} & \best{40.98} & \best{40.58} & \best{40.38} & \best{40.08} \\

HOI-DA (Swin-T~\cite{liu2021swin}) & Video(+Text)
& 27.69 & 27.52 & 27.08 & 26.53 & 26.20
& 33.68 & 33.97 & 33.65 & 33.06 & 32.67
& 38.52 & 39.29 & 39.16 & 38.74 & 38.53 \\
\bottomrule
\end{tabular}%
}
\end{table*}

\section{Supplementary Ablation Results}
\label{sec:supp-ablation}
 
\noindent\textbf{Recall@$k$ metric.}
We supplement the mAP-based ablation in Section~5.5 of the main paper with
Recall@$k$ on the VidHOI component of DETAnt-HOI.
We follow the person-wise top-$k$ evaluation protocol used in
Gaze-Tran~\cite{ni2023gaze}, which extends frame-level recall to
multi-person scenarios.
Predicted HOI triplets are first assigned to person identities via
bounding-box IoU matching.
Within each person, predictions are ranked by HOI confidence score and the
top-$k$ are retained; each is classified as a true or false positive, and
a per-person recall is computed.
The reported Recall@$k$ is the mean over all annotated persons in the
evaluation set.
 
\noindent\textbf{Interpreting metrics across horizons.}
As the anticipation horizon increases, ground-truth labels become sparser
and long-tail interaction categories tend to disappear from the supervision
first.
Because mAP is averaged over the active category set, this shrinkage can
inflate mAP at longer horizons as rare categories vanish.
Recall@$k$ is computed per person independently of the category vocabulary
and is therefore immune to this effect.
We report both metrics to provide a complete and unbiased picture of how
each design choice affects present-time grounding and future anticipation.
 
\noindent\textbf{Ablation results (Table~\ref{tab:supp-ablation}).}

The recall-based ablations are broadly consistent with the mAP results in Table~5 of the main paper, in the sense that the naively shared unified one-decoder design remains the weakest configuration, while the variants that preserve explicit temporal or semantic structure maintain substantially stronger ranking of plausible HOIs over the same pair hypotheses. This trend is especially clear at Recall@20 and Recall@50, where the gap between the unified one-decoder baseline and the stronger structured variants persists from \(h=0\) to \(h=7\). At the same time, the exact ordering is not identical to mAP, which is expected: Recall@k measures whether the correct HOI remains within the top-\(k\) predictions, whereas mAP is more sensitive to confidence calibration and false-positive suppression.

Removing the Temporal Summary Module leads to a consistent reduction across Recall@10, Recall@20, and Recall@50, both at the observation boundary and across future horizons. This confirms that future anticipation benefits from conditioning on the observed interaction history rather than on the boundary frame alone. Notably, the degradation is not concentrated exclusively at \(h=7\); instead, it is distributed across both near-term and longer-term prediction, suggesting that temporal summarization stabilizes pair-level ranking throughout the forecasting window rather than only at the farthest horizon.

The language-related ablations reveal a clearer separation at higher-recall regimes. Removing decoder-side text guidance does not induce the dominant degradation and remains relatively close to the full model, whereas removing the language-guided semantic branch causes a substantially larger and more consistent drop, especially under Recall@50. This pattern indicates that language contributes primarily by regularizing the shared interaction space and improving the ranking of semantically plausible future HOIs under a larger candidate set, while decoder-side guidance provides a more localized gain.

By contrast, the effect of dual orthogonality regularization is less pronounced in Recall@k than in mAP. Its contribution is not strictly monotonic with horizon and does not appear as a uniform gain at longer forecasting steps. This suggests that the regularizer mainly improves the separation and calibration of present grounding, future change, and horizon-specific dynamics, effects that are reflected more directly in mAP than in top-\(k\) recall. Taken together, Table~\ref{tab:supp-ablation} supports the same overall conclusion as the main paper: the benefit of HOI-DA does not come from naive task fusion, but from structuring the shared pair-centric representation so that present grounding and future anticipation reinforce each other over time.

\begin{table*}[t!]
\centering
\caption{%
  \textbf{Recall@$k$ results for the ablation study on the
  VidHOI component of DETAnt-HOI.}
  $h{=}0$ denotes present-time detection; $h{>}0$ denotes future
  anticipation.
  The mAP counterpart of this table appears in Table~5 of the main paper.
}
\vskip-2ex
\label{tab:supp-ablation}
\small
\setlength{\tabcolsep}{4pt}
\renewcommand{\arraystretch}{1.10}
\resizebox{\linewidth}{!}{%
\begin{tabular}{lccccc|ccccc|ccccc}
\toprule
\multirow{2}{*}{\textbf{Method}} &
\multicolumn{5}{c|}{\textbf{Recall@10} $\uparrow$} &
\multicolumn{5}{c|}{\textbf{Recall@20} $\uparrow$} &
\multicolumn{5}{c}{\textbf{Recall@50} $\uparrow$} \\
\cmidrule(lr){2-6}\cmidrule(lr){7-11}\cmidrule(lr){12-16}
&
\textbf{$h{=}0$} & \textbf{$h{=}1$} & \textbf{$h{=}3$} & \textbf{$h{=}5$} & \textbf{$h{=}7$} &
\textbf{$h{=}0$} & \textbf{$h{=}1$} & \textbf{$h{=}3$} & \textbf{$h{=}5$} & \textbf{$h{=}7$} &
\textbf{$h{=}0$} & \textbf{$h{=}1$} & \textbf{$h{=}3$} & \textbf{$h{=}5$} & \textbf{$h{=}7$} \\
\midrule
Separate Det.\ \& Ant. & 49.30 & 49.65 & 51.04 & 51.23 & 50.40 & 57.53 & 58.03 & 59.19 & 59.38 & 59.52 & 60.80 & 61.22 & 61.93 & 62.32 & 62.76 \\
Unified One-Decoder & 48.69 & 46.33 & 47.89 & 48.83 & 49.52 & 54.28 & 54.62 & 56.01 & 57.02 & 57.54 & 58.72 & 58.73 & 59.63 & 60.51 & 61.01 \\
w/o Temporal Summary Module & 55.97 & 56.78 & 57.66 & 58.36 & 58.89 & 59.65 & 60.35 & 61.21 & 61.82 & 62.28 & 61.11 & 61.82 & 62.59 & 63.10 & 63.59 \\
w/o Decoder-Side Text Guidance & 56.37 & 57.18 & 58.06 & 58.76 & 59.29 & 60.05 & 60.75 & 61.61 & 62.22 & 62.68 & 61.51 & 62.22 & 62.99 & 63.50 & 63.99 \\
w/o Lang.-Guided Semantic Branch & 54.02 & 54.32 & 55.52 & 56.01 & 56.65 & 57.54 & 57.83 & 58.89 & 59.55 & 60.14 & 58.82 & 59.79 & 60.13 & 60.70 & 61.31 \\
w/o Dual Orthogonality Reg. & 57.87 & 58.15 & 59.24 & 59.90 & 60.29 & 61.29 & 61.53 & 62.41 & 63.02 & 63.42 & 62.69 & 63.05 & 63.85 & 65.20 & 64.77 \\
\best{HOI-DA (full)}
& \best{57.54} & \best{58.08} & \best{59.01} & \best{59.60} & \best{59.92}
& \best{61.50} & \best{62.08} & \best{62.98} & \best{63.56} & \best{64.03}
& \best{63.29} & \best{63.97} & \best{64.73} & \best{65.25} & \best{65.68} \\
\hline
\end{tabular}
}
\end{table*}
 
\section{Additional Qualitative Results}
\label{sec:supp-qual}
 
We provide three complementary sets of qualitative results, each targeting
a distinct aspect of HOI-DA's behavior.
 
\begin{figure*}[t!]
  \centering
  \includegraphics[width=\textwidth]{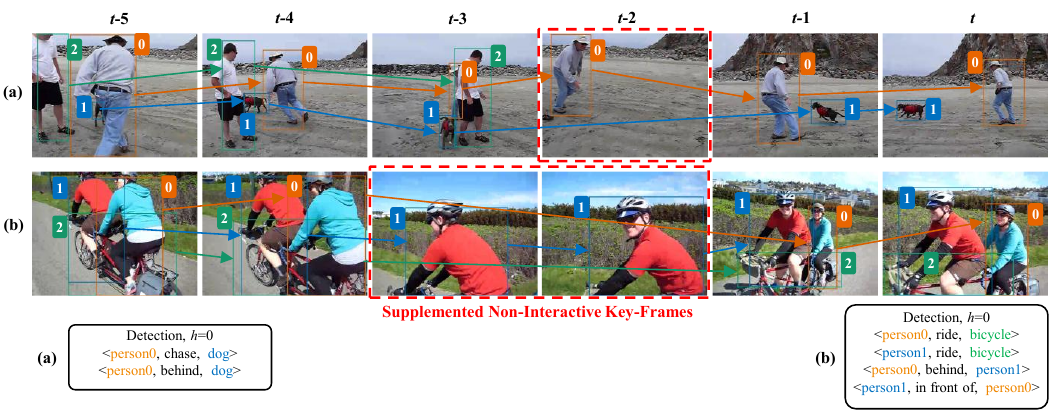}
  \vskip-1ex
  \caption{%
    \textbf{Pair tracking under occlusion and camera motion on VidHOI.}
    Numbered bounding boxes denote persistent human--object pair hypotheses
    maintained by HOI-DA across the full observation window.
    \textit{Top:} a person and an object temporarily exit the field of view
    due to instance motion.
    \textit{Bottom:} rapid camera movement causes severe occlusion and motion
    blur.
    Red-bordered frames are supplementary non-interactive key-frames
    introduced by DETAnt-HOI to preserve instance positional continuity.
    Despite disappearance and occlusion, HOI-DA correctly maintains pair
    identity and resumes accurate interaction prediction upon reappearance,
    in contrast to two-stage pipelines that must re-identify instances from
    scratch.
  }
  \Description{%
    Two video scenarios showing temporary instance disappearance and
    occlusion. Numbered bounding boxes track persistent human--object pairs.
    Red-bordered frames mark supplementary non-interactive key-frames from
    DETAnt-HOI. HOI-DA maintains correct pair identity and interaction
    predictions throughout both scenarios.
  }
  \label{fig:track}
\end{figure*}

\noindent\textbf{Pair tracking under occlusion and camera motion
(Figure~\ref{fig:track}).}
Figure~\ref{fig:track} examines two challenging scenarios: one where a person
and an object temporarily exit the field of view due to instance motion
(top), and one where rapid camera movement causes partial occlusion and
motion blur (bottom).
Two-stage pipelines that depend on frame-by-frame detector outputs face the
re-identification problem in both cases. Once an instance is missed in a
single frame, the tracker must decide whether the next detection is the same
entity or a new one, and errors compound over time under significant motion
or occlusion.
HOI-DA sidesteps this by maintaining pair identity in persistent slot
queries that attend over the full spatio-temporal memory; no per-frame
continuity is required.
In the top scenario, person-2 and the dog leave the frame at $t{-}2$ while
person-0 is heavily occluded at $t{-}3$; the model nevertheless resumes
correct tracking upon their reappearance.
Frames with red borders correspond to supplementary non-interactive
key-frames introduced by DETAnt-HOI.
Although these frames carry no HOI labels, the positional annotations they
provide ensure that bounding-box trajectories remain continuous across short
inactive gaps, preventing the spatial discontinuities that would otherwise
disrupt pair slot attention.
 
\begin{figure*}[t!]
  \centering
  \includegraphics[width=\textwidth]{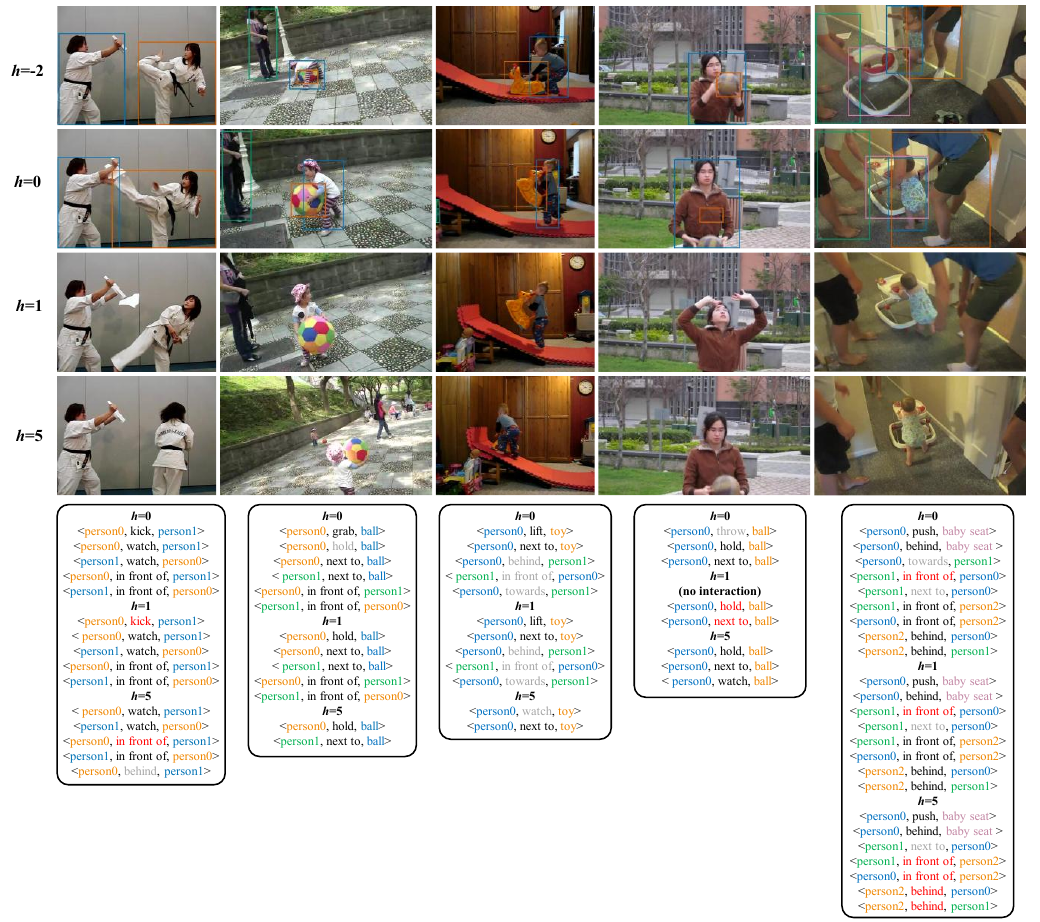}
  \vskip-1ex
  \caption{%
    \textbf{Joint HOI detection and multi-horizon anticipation across diverse
    scenes on VidHOI.}
    For each scenario, we visualize predictions for persistent human--object pair hypotheses at multiple temporal offsets, including an observed past frame at $h{=}{-}2$, present-time detection at $h{=}0$, and future anticipation at $h{=}1$ and $h{=}5$. Here, $h{=}{-}2$ denotes the frame two steps before the current observation boundary within the six-frame input window.
    The model captures systematic temporal transitions such as
    \textit{grab}$\to$\textit{hold} and \textit{lift}$\to$\textit{hold},
    while errors concentrate on sustained or cyclic interactions where
    continuation duration is ambiguous.
    Both successful and failed predictions are shown.
    Verb colors: black (true positives), {\color{red}red} (false positives),
    {\color{gray}gray} (false negatives).
  }
  \Description{%
    Multiple video scenarios with HOI detection and anticipation results at
    three horizons. Each row tracks a persistent human--object pair. True
    positives are in black, false positives in red, false negatives in gray.
  }
  \label{fig:visual2}
\end{figure*}

\noindent\textbf{Multi-scenario detection and anticipation.}
Figure~\ref{fig:visual2} presents joint detection and anticipation results
across diverse scenes and interaction types, displaying predictions at
$h{=}0$ (present detection), $h{=}1$ (short-term anticipation), and
$h{=}5$ (long-term anticipation).
The model consistently identifies primary action-type relations and captures
temporal transitions such as \textit{grab}$\to$\textit{hold} and
\textit{lift}$\to$\textit{hold}, reflecting common physical interaction
sequences.
Errors concentrate on sustained or cyclic interactions such as \textit{kick},
where the duration of continuation is inherently ambiguous and annotation
noise further complicates supervision.
We include both successful and failed predictions to provide an honest
characterization of where the unified formulation helps and where open
challenges remain.
 
\begin{figure*}[t!]
  \centering
  \includegraphics[width=\textwidth]{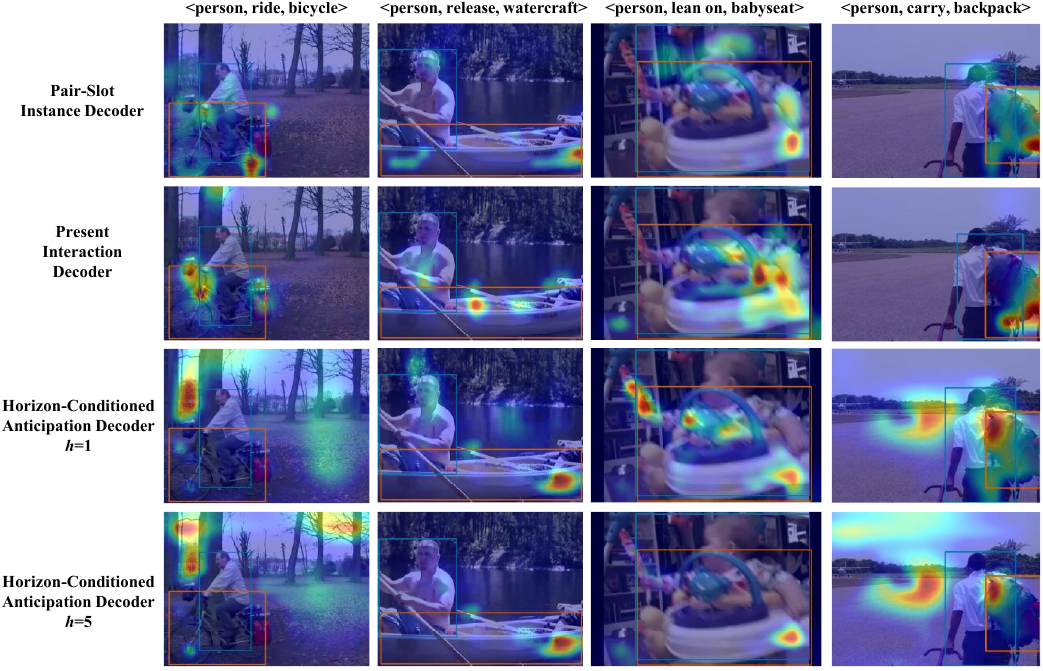}
  \vskip-1ex
  \caption{%
    \textbf{Cross-attention heatmaps at three decoder stages of HOI-DA on
    VidHOI.}
    \textit{Top (instance localization):} attention concentrates on
    texture-discriminative regions, including faces and limbs for humans and
    shape boundaries for objects.
    \textit{Center (present interaction detection):} attention shifts to the
    hand--object contact zone, which carries the strongest signal for
    fine-grained predicate classification.
    \textit{Bottom (future anticipation):} attention broadens to incorporate
    gaze direction and environmental context beyond the current contact zone.
    This progression is consistent with the residual formulation in Eq.~(6)
    of the main paper, where the localization and detection stages anchor the
    present pair state and the anticipation stage attends to cues that predict
    how the interaction will evolve.
  }
  \Description{%
    Three columns of cross-attention heatmaps for the localization,
    detection, and anticipation decoder stages. Attention progresses from
    instance-level texture regions, to hand-object contact zones, to broader
    gaze and environmental context.
  }
  \label{fig:attn}
\end{figure*}

\noindent\textbf{Decoder attention heatmaps (Figure~\ref{fig:attn}).}
Figure~\ref{fig:attn} visualizes cross-attention weights at three stages of
the unified pair-centric decoder: instance localization, present interaction
detection, and future anticipation.
At the localization stage (left), attention concentrates on
texture-discriminative regions of the relevant instances, such as facial and
limb areas for humans and characteristic shape boundaries for objects.
At the detection stage (center), attention shifts toward the contact zone
between human and object, such as the hand--object interface, which
carries the strongest signal for fine-grained predicate classification.
At the anticipation stage (right), attention broadens to incorporate gaze
direction and environmental context beyond the current contact zone.
This progressive widening is consistent with the residual formulation in
Eq.~(6) of the main paper. The localization and detection stages anchor
the pair representation in its present state, while the anticipation stage
attends to the contextual cues that predict how the interaction will evolve.
Task-specific embeddings
($\mathbf{e}^{\mathrm{det}}_{\mathrm{task}}$ and
$\mathbf{e}^{\mathrm{ant}}_{\mathrm{task}}$)
decouple these attention patterns across decoder stages, allowing each to
specialize without interfering with the others.
 
\end{document}